\documentclass[10pt,twocolumn,letterpaper]{article}

\usepackage{cvpr}
\usepackage{times}
\usepackage{epsfig}
\usepackage{graphicx}
\usepackage{amsmath}
\usepackage{amssymb}

\usepackage[breaklinks=true,bookmarks=false]{hyperref}
\usepackage[ruled,vlined]{algorithm2e}
\usepackage{makecell}

\cvprfinalcopy 

\def\cvprPaperID{****} 

\begin{document}

\title{Semi-Supervised Object Detection with Sparsely Annotated Dataset}

\author{
Jihun Yoon\\
hutom\\
Republic of Korea\\
{\tt\small yjh2020@hutom.io}
\and
Seungbum Hong\\
hutom\\
Republic of Korea\\
{\tt\small qbration21@hutom.io}
\and
Sanha Jeong\\
hutom\\
Republic of Korea\\
{\tt\small jeongsanha@hutom.io}
\and
Min-Kook Choi\\
hutom\\
Republic of Korea\\
{\tt\small mkchoi@hutom.io}
}

\maketitle

\begin{abstract}
In training object detector based on convolutional neural networks, selection of effective positive examples for training is an important factor. However, when training an anchor-based detectors with sparse annotations on an image, effort to find effective positive examples can hinder training performance. When using the anchor-based training for the ground truth bounding box to collect positive examples under given IoU, it is often possible to include objects from other classes in the current training class, or objects that are needed to be trained can only be sampled as negative examples. We used two approaches to solve this problem: 1) the use of an anchorless object detector and 2) a semi-supervised learning-based object detection using a single object tracker. The proposed technique performs single object tracking by using the sparsely annotated bounding box as an anchor in the temporal domain for successive frames. From the tracking results, dense annotations for training images were generated in an automated manner and used for training the object detector. We applied the proposed single object tracking-based semi-supervised learning to the Epic-Kitchens dataset. As a result, we were able to achieve \textbf{runner-up} performance in the \textbf{Unseen} section while achieving the \textbf{first place} in the \textbf{Seen} section of the Epic-Kitchens 2020 object detection challenge under IoU $>0.5$ evaluation.
\end{abstract}

\vspace{-5mm}

\section{Introduction}

Thanks to the rapid development of CNN (Convolutional Neural Networks), the performance of object recognition networks using CNN has also been improved dramatically \cite{Li18}. As the performance of the object detection network has been improved, the dataset for evaluating it was also started from a dataset with low complexity such as PASCAL VOC \cite{Mark15} and developed to have a high complexity such as MS-COCO \cite{TsungYi14}. Among the object detection datasets, the relatively recently released Epic-Kitchens dataset has the following characteristics different from other object detection datasets \cite{Dima18}.

\begin{itemize}
  \item Images for training detector are collected from the original video, and corresponding frame sequences are provided.  \vspace{-1mm}
  \item In a training image, only some of the trainable objects are sparsely annotated. \vspace{-1mm}
  \item The difference in the amount of annotations between the few and many shot classes is large, depending on the distribution of the appearance of the objects in the training dataset. \vspace{-1mm}
\end{itemize}

\begin{figure}[t!]
\centering
\includegraphics[width=0.7\linewidth]{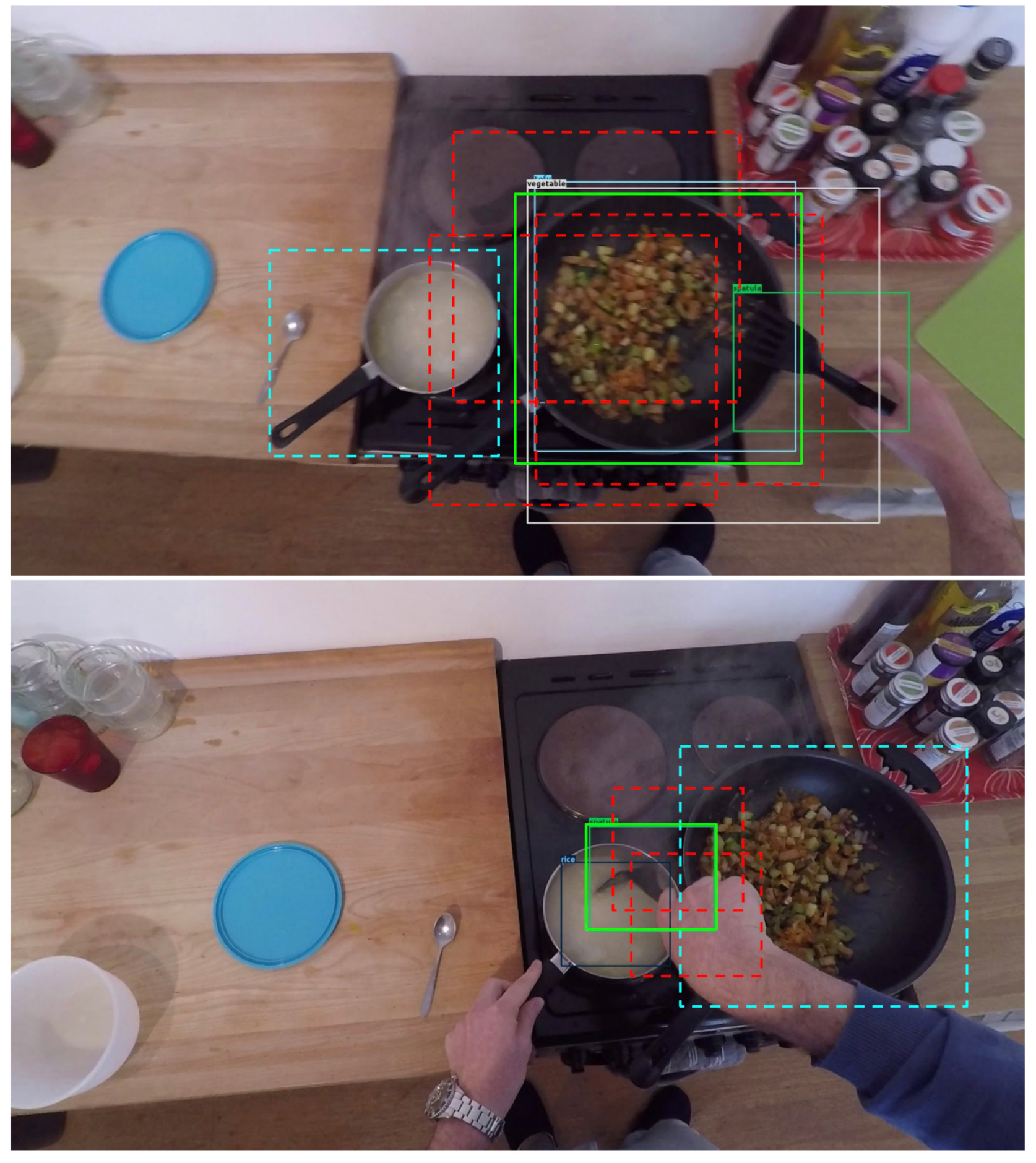}
\caption{\textbf{An example of anchor-based detector training on a sparsely annotated dataset.} The solid green line represents the label information in each training image, and the red dotted line is an example of positive examples. Light blue dashed lines indicate objects that are included in the label in other training images (top), but are not labeled in the given image to train (bottom). As such, in the Epic-Kitchens object detection dataset, it is an object to learn when training an anchor-based detector, but training performance is impaired because label information is missing.}
\label{fig:example}
\vspace{-5mm}
\end{figure}

As described above, the annotation of Epic-Kitchens for object detection is provided in a different way from the existing dataset, and has a characteristic that it is difficult to apply the method of training the existing object detection model as it is. Typically, in the case of detectors that train positive examples based on anchors \cite{Joseph17,Wei16} or detectors that train the objectness of a candidate object with the structure of an RPN \cite{Shaoqing15, Jifeng18}, batch sampling is performed considering the Intersection on Union (IoU) with the ground truth bounding box for effective training. However, if an anchor-based hard example mining is performed on a sparsly annotated training image, the efficiency of training is hindered by the distribution of objects near the ground truth bounding box. Figure 1 shows the negative effects of trying to select a good positive example when training anchor-based object detectors with sparse annotation images.

We have trained the object detector through two approaches to solve this problem. First, a detector using different parameterization for estimating the bounding box was used instead of the anchor-based detection model. We were able to minimize the effect of sparsely annotated training images affected by anchor-based sampling by utilizing the Fully Convolutional One-Stage Object Detection (FCOS) network \cite{Zhi19}. The second is to utilize the features of the Epic-Kitchens dataset, and the object tracking technique is used to semi-supervise all detectable objects in each training image. To this end, the bounding box label existing in a specific frame is set as an initial bounding box on the time domain and used as an input of a single object tracker. Subsequently, the predictive output exceeding the threshold from the tracker was assumed to be a pesudo annotation, and labels for all learnable objects were provided in the image for training. Figure 2 schematically shows the proposed goal of semi-supervised learning.

With the proposed approach, we were able to train an object detection network effectively with the Epic-Kitchens object detection dataset. Subsequently, joint NMS-based ensemble \cite{Heechul17} was performed for FCOS models with trained inhomogeneous backbones. As a result, we were able to achieve \textbf{\textit{first place}} in the \textit{Seen} evaluation set and \textbf{\textit{runnerup}} in the \textit{Unseen} evaluation set under the IoU $>0.5$ of the Epic-Ktichens 2020 object detection challenge.

\begin{figure}[t!]
\centering
\includegraphics[width=0.9\linewidth]{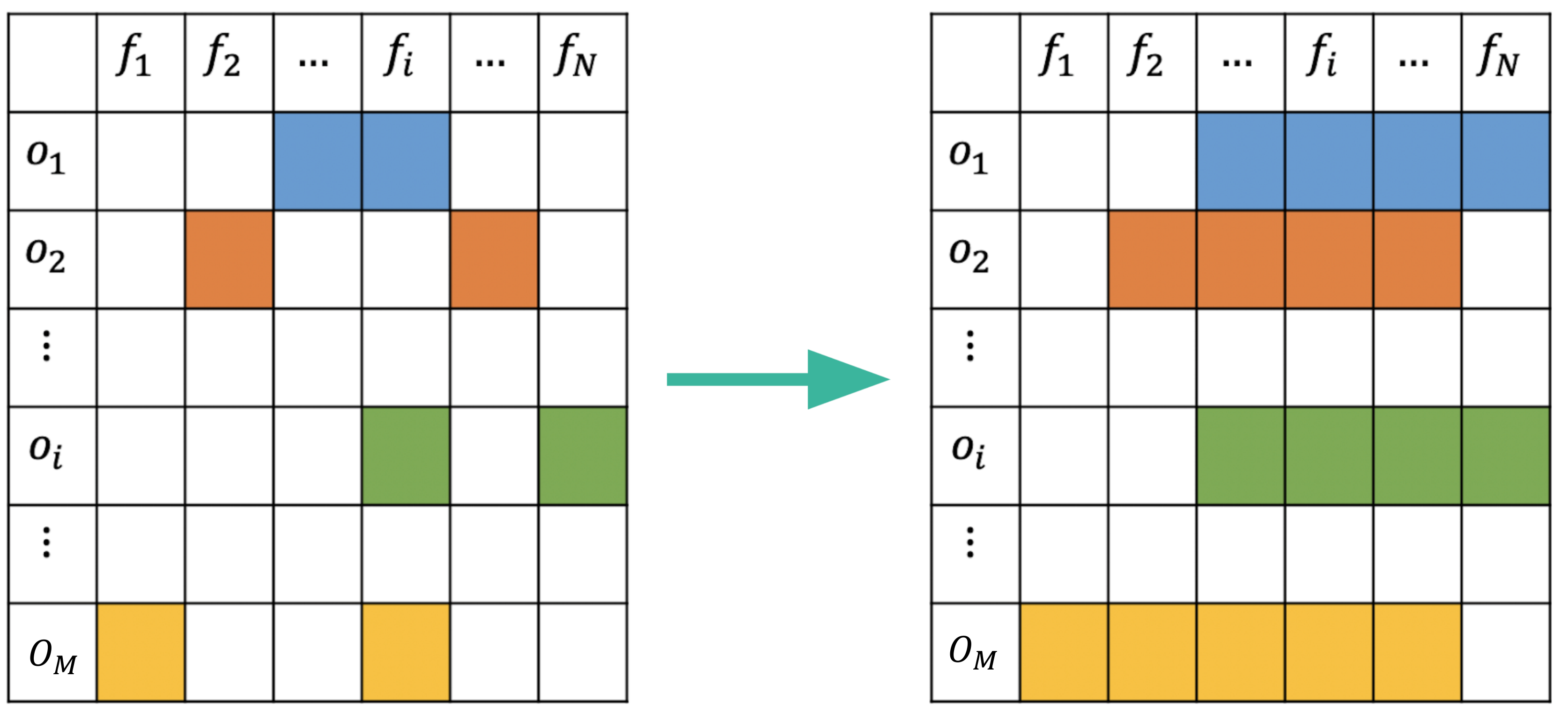}
\caption{\textbf{The final goal of the proposed supervised learning.} $f$ denotes an action clip composed of $N$ frames, and $o$ denotes a total of $M$ objects present in the action clip. We performed semi-supervised learning through bidirectional tracking to obtain dense labels for all learnable objects present in the action clip.}
\label{fig:example}
\vspace{-5mm}
\end{figure}

\section{Related Work}

\noindent \textbf{Object detection.}  CNN-based object detection models are largely divided into one-stage or two-stage models. In the one-stage model, the process of predicting the class and position of an object is performed in one structure, and examples include YOLO \cite{Joseph17,Joseph18}, SSD \cite{Wei16,ChengYang17}, and RetinaNet \cite{TsungYi17}. Additional structures such as Feature Pyramid Networks (FPN) \cite{Joseph17} are often used to efficiently process the output features obtained from the backbone in the head structure. In general, it is known that the regression accuracy is lower than the two-stage model because classification and regression are performed in one structure. In the case of the two-stage model, the prior knowledge of the location of the object is estimated from the RPN (Region Proposal Network), which is a subnetwork in the detector \cite{Shaoqing15}. RPN determines objectness by class-agnostic subnet, and performs class-aware detection through the subsequent head structure. Faster R-CNN \cite{Shaoqing15}, R-FCN \cite{Jifeng18}, Cascade R-CNN \cite{Zhaowei18}, Cascade RPN \cite{Thang19}, etc. are representative of various head structures, and are known to have relatively high regression accuracy. Models such as RefineDet \cite{Shifeng18} that combine the philosophy of one-stage and two-stage models have also been proposed, and detectors that utilize other parameterizations for bounding box regression rather than structural advantages, such as FCOS \cite{Zhi19}, have also been proposed. \\

\noindent \textbf{Semi-supervised learning for object detection.}  Object detection using semi-supervised learning is used in situations where it is difficult to manually acquire a sufficient number of annotations to learn, or when pseudo labels are to be obtained from a relatively large number of unlabeled data. \cite{Ishan15} proposed an iterative framework for evaluating and retraining pseudo labels using pre-trained object detectors and robust trackers to obtain good pseudo labels in successive frames. In \cite{Yusuke19}, it was possible to achieve improved detection performance in Open Image Dataset V4 by utilizing part-aware sampling and RoI proposals to obtain good pseudo labels for sparsely annotated large-scale datasets. In \cite{MinKook18}, in order to efficiently use unlabeled data from the MS-COCO dataset, co-occurrent matrix analysis was used to generate good pseudo labels by using prior information of the labeled dataset. The proposed single object tracker-based semi-supervised learning is similar to \cite{Ishan15} in that it uses a tracker, but has a difference in obtaining dense annotation information for a specific image by using the existing lean annotation information. At the same time, since the object detector is not used for the initial input for tracking, the training is not applied as an iterative training scenario. \\

\noindent \textbf{Single object visual tracking.}  We used a single object tracking network to generate pseudo labels for sparsely annotated datasets. In single object tracking \cite{Luca16,Bo18,Qiang19,Bao19,Bo19}, the Siamese network-based visual tracker shows balanced accuracy and speed across various datasets. The Siamese network-based tracker basically trains with the similarity of the CNN feature for the target image and the input image for tracking. We used SiamMask \cite{Qiang19} as a single object tracker, which uses box and mask information together with similarity of features to the tracking target.
\section{Fully Convolutional One-Stage Object Detection (FCOS)}

We used the FCOS model \cite{Zhi19} to exclude the computational process for selecting a good positive example with an anchor from detector training. FCOS defined the parameters for regression of the bounding box differently, and presented an anchor-free detector. The loss function of FCOS is defined as follows.

\begin{align}
\begin{split}
  L(a_p,m_p)=\frac{1}{N}\sum_{\{u,v\} \in p}L_{cls}(a_p, c_p)+\\
  \frac{1}{N}\sum_{\{u,v\} \in p}[c_p > 0 ]L_{reg}(m_p, \hat{m}_p),
\end{split}  
\end{align}

\noindent where $p$ denotes a position $(u,v)$ on the feature map, and $a_p$ denotes a prediction vector for class estimation. $c_p$ denotes the class label for the input example, and $m_p$ denotes the specific spatial location of the feature map and the distance from the ground truth bounding box $m_p=(l,t,r,b)$. Given the label $B=(x_0,y_0,x_1,y_1,c)$ for the bounding box, FCOS parameterizes to find the bounding box by $l=u-x_0$, $t=v-y_0$, $r=u-x_1$, $b=v-y_1$. We used pretrained ResNet \cite{Kaiming16}, ResNeXt \cite{Saining17}, HRNet \cite{Jingdong19} as the backbone network for FCOS to perform inhomogeneous ensemble.

\section{Semi-Supervised Learning with Single Object Tracker}

The Epic-Kitchens dataset simultaneously provides a bounding box label for a particular object and a sequence frame for the \textit{action clip} in which the object appears. The bounding box for the object is not densely given every frame, but rather sparsely in the action sequence. We used a single object tracker to achieve the goal in Figure 2 with an automated procedure. Among various single object trackers, SiamMask \cite{Qiang19} was used, which shows a balanced performance for tracking accuracy and speed. We performed a bidirectional tracking using the SiamMask model trained from the DAVIS dataset, using each bounding box as the initial value for a single object. The details of forward tracking with SiamMask for one action clip input are described in Algorithm 1. Algorithm 1 is used in the same way for backward tracking to complete bidirectional tracking. Figure 3 shows an example of the start and end of tracking according to Algorithm 1 on a single object, and Figure 4 shows training images with pseudo labels generated after tracking an object.

\begin{algorithm}[t!]
\SetAlgoLined
 Input: Action clip ($A$), pretrained tracking model ($T$), a set of bbox for initial input ($BB$), threshold of tracking score ($\rho_1$), threshold of IoU between two pair of tracked bbox ($\rho_2$)\\
 Output: $BB$ in $Q$ from $T$ \\
 Initialize an empty queue $Q$ \\
 \While{bbox $b_{c,i}$ with class $c$ at $i$-th frame available from $BB$}{
  Get a list of frames $FF$ in forward from $i$-th frame in $A$; \\
  Initialize $T$ with $b_i$ from $A$; \\
  Initialize a variable $pre_s$ with a size of $b_{c,i}$ to store a size of object from $T$ at previous frame; \\
  \While{each frame in FF}{
  Get a bbox $b_{c,k}$ from $T$ at $k$-th frame; \\
  $crnt_s$ $:=$ a size of $b_{c,k}$; \\
  \eIf{$IoU(prev_s,crnt_s) \geq\rho_1$  }{
   $prev_s := crnt_s$; \\
   Add $b_{c,k}$ to $Q$; \\
   }{
   break; \\
  }
 }
 }
 \caption{Forward tracking}
\end{algorithm}

\begin{figure}[t!]
\centering
\includegraphics[width=1.0\linewidth]{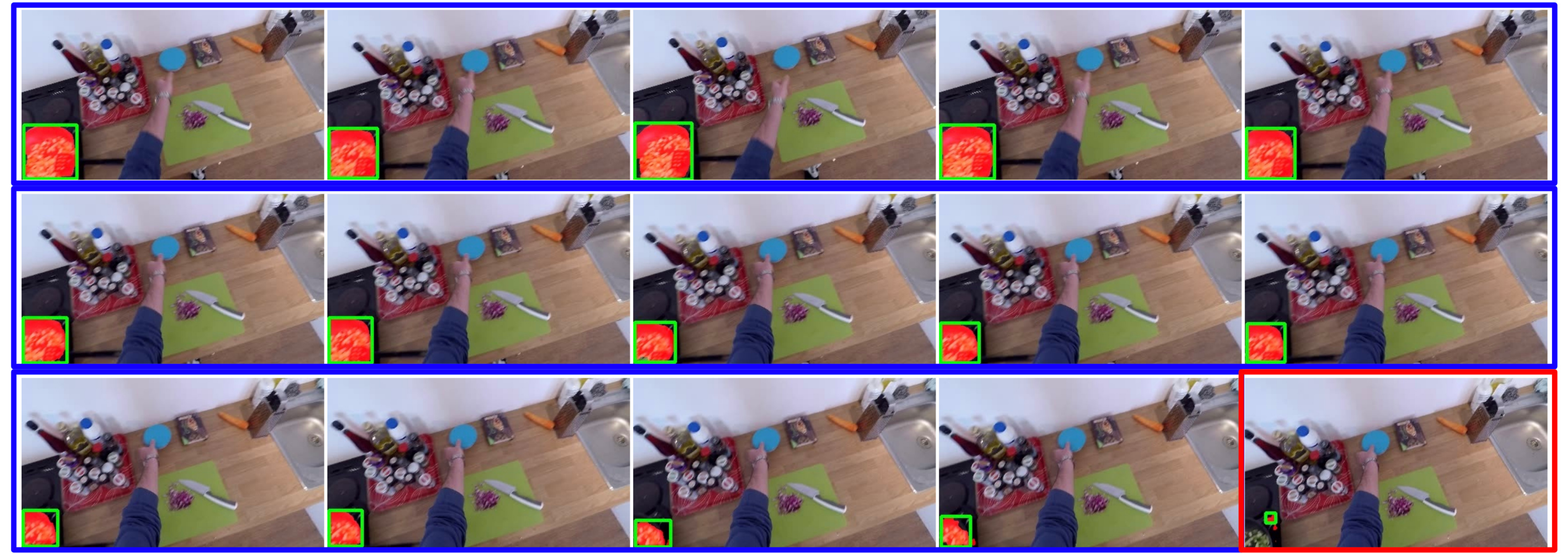}
\caption{\textbf{Example of the result of Algorithm 1.} Frames marked with blue boxes are frames that have been tracked with the same object since tracking started, and a frame marked with a red box is a frame whose tracking is terminated due to the termination condition of algorithm 1.}
\label{fig:example}
\end{figure}

\begin{figure}[t!]
\centering
\includegraphics[width=1.0\linewidth]{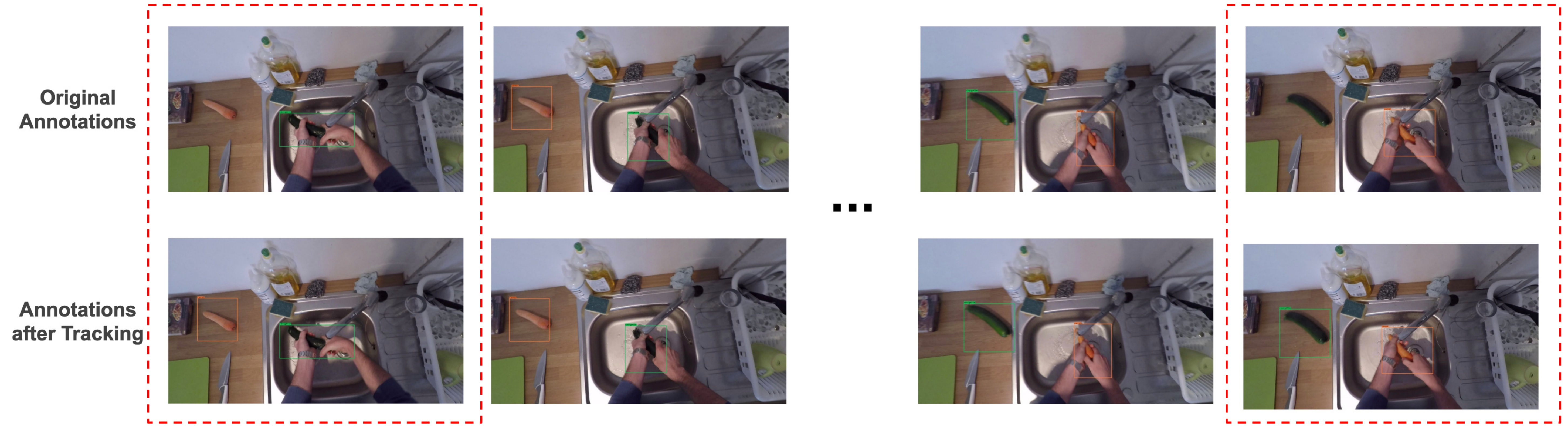}
\caption{\textbf{Changes in training images after tracking.} An example of the final annotations to be used for semi-supervised learning is shown on the training images indicated by the red dotted line.}
\label{fig:example}
\end{figure}

\section{Epic-Kitchens Object Detection Results}

\noindent \textbf{Training details.} We used Faster R-CNN \cite{Shaoqing15} and Cascade R-CNN \cite{Zhaowei18} as anchor-based detectors and FCOS \cite{Zhi19} as an anchorless detector to compare performance in the Epic-Kitchens object detection dataset \cite{Dima18}. As the backbone CNN for training the detector, ResNet-50, ResNet-101, ResNeXt-101, and HRNet-V2p-W32 pretrained with ImageNet were used, and training details for each combination of backbone and head structure are shown in Table 1. All experiments were conducted using the MMDetection library \cite{Kai19}. \\

\begin{table}[t!]
  \caption{\textbf{Training detail of each detector.} `lr' represents the learning rate. Learning schedule is indicated as (scheduler, drop rate) [drop epoch1:drop epoch2:max epoch].}
  \label{tab:freq}
  \resizebox{\linewidth}{!}{
  \begin{tabular}{c|c|c|c|c}
\hline
 Detector & Backbone & Optimizer (lr) & Learning schedule & Warmup(iter, ratio) \\ \hline
Faster R-CNN & ResNet-101 & SGD (0.02) & (step, 0.1)[8:11:24]& linear(500, $1/3$)\\ 
Cascade R-CNN & HRNet-V2P & SGD (0.02) & (step, 0.1)[16:19:30]& linear(500, $1/3$)\\ 
FCOS & HRNet-V2P & SGD (0.01) & (step, 0.1)[8:11:20]& constant(500, $1/3$)\\
FCOS & ResNet-50 & SGD (0.01) & (step, 0.1)[8:11:16]& constant(500, $1/3$)\\ 
FCOS & ResNet-101 & SGD (0.01) & (step, 0.1)[16:20:22]& constant(500, $1/3$)\\ 
FCOS & ResNeXt-101 & SGD (0.01) & (step, 0.1)[16:22:39]& constant(500, $1/3$)\\ 
\hline
\end{tabular}
}
\end{table}

\noindent \textbf{Anchor-based vs. anchorless detector} Table 2 shows the training performance in a single model of an anchor-based detector and an anchorless detector. According to Table 2, it can be seen that in the basic performance of training, the performance of the detector without an anchor is excellent and shows stable learning results. Figure 5 shows the loss change during training of an anchor-based detector and an anchorless detector. The detector without anchor shows a relatively stable loss curve. At the same time, Table 2 shows the performance change of the FCOS model according to different backbones. For a single model, it was confirmed that the FCOS model utilizing the ResNet-101 backbone achieved the best generalization performance in the Seen set, and the HRNet backbone model performed the best in the Unseen set. \\

\begin{table}[t!]
  \caption{\textbf{Performance comparison of anchor-based and anchorless detectors.} If the model name has a `$+$', it is the result of evaluation using tracker-based semi-supervised learning. The highest performance in a single model and the highest performance in an entire model are shown in bold.}
  \label{tab:freq}
  \resizebox{\linewidth}{!}{
  \begin{tabular}{c|c|c|c|c|c|c|c}
\hline 																
Detector & backbone & \multicolumn{3}{|c|}{Seen} & \multicolumn{3}{|c}{Unseen} \\ \hline
- & - & $>$ 0.05 & $>$ 0.5 & $>$ 0.75 & $>$ 0.05 & $>$ 0.5 & $>$0.75 \\ \hline
Faster R-CNN & ResNet-101 & 37.54 & 28.64 & 6.92 & 32.83 & 23.16 & 5.55 \\  
Cascade R-CNN & HRNet-V2P & 30.44 & 24.17 & 8.73 & 23.87 & 18.05 & 6.81 \\  
FCOS & HRNet-V2P & 48.44 & 34.87 & \textbf{11.02} & \textbf{43.88} & \textbf{30.68} & \textbf{9.27} \\  
FCOS & ResNet-50 & 46.96 & 34.51 & 10.09 & 42.46 & 29.49 & 7.48 \\  
FCOS & ResNet-101 & 49.77 & 35.8 & 10.15 & 43.39 & 28.98 & 7.86 \\  
FCOS & ResNeXt-101 & 48.17 & 33.95 & 9.86 & 41.79 & 27.27 & 7.19 \\  
FCOS+ & ResNet-101 & \textbf{50.27} & \textbf{35.89} & 10.57 & 43.14 & 29.82 & 7.76 \\ 
FCOS Ensemble+ & - & \textbf{58.27} & \textbf{44.48} & \textbf{15.36} & \textbf{55.72} & \textbf{41.12} & \textbf{12.5} \\  
\hline
\end{tabular}
}
\end{table}

\begin{figure*}[t!]
\centering
\includegraphics[width=1.0\linewidth]{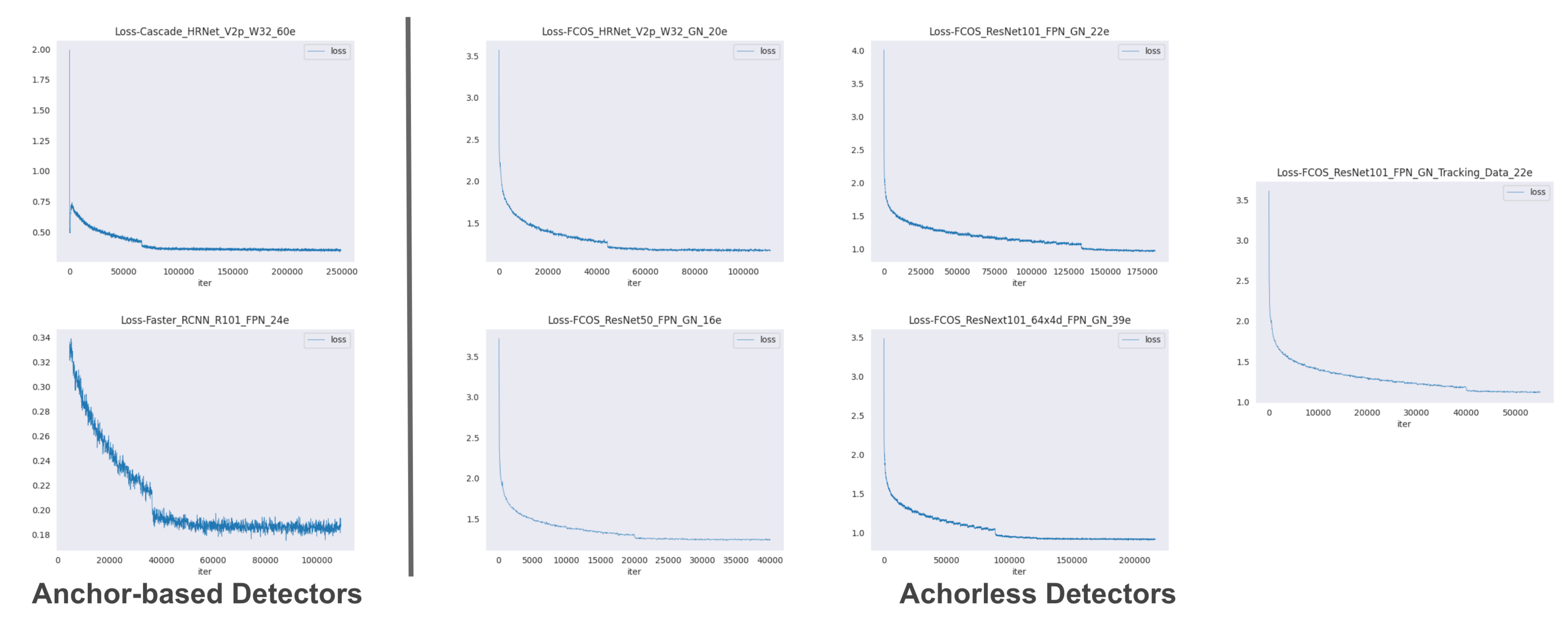}
\caption{\textbf{Training losses for anchor-based and anchorless detectors.} It shows that the loss curve of the anchorless detectors is relatively stable.}
\label{fig:example}
\end{figure*}

\noindent \textbf{Semi-supervised learning.}  We used the pretrained SiamMask model from the DAVIS dataset \cite{Jordi17} to generate dense labels to train the FCOS models with dense annotations. Table 2 shows that the generalization performance of the FCOS models under $IoU>0.5$ is consistently improved when using semi-supervised learning based on single object tracking. \\

\noindent \textbf{Inhomogeneous backbone ensemble.} We performed model ensemble for each trained model in Table 2 to achieve the best detection performance from trained detectors. We performed the ensemble using the joint NMS technique \cite{Heechul17}, where we can achieve an ensemble by applying NMS to a bounding box with up to 300 high prediction scores obtained from every detectors. Table 2 shows the performance change of the Seen and Unseen sets according to the ensemble combination, and Figure 6 shows the performance published on the Epic-Kitchen object detection challenge page. We were finally able to achieve the \textbf{\textit{first rank}} in \textit{Seen} set and \textbf{\textit{runnerup}} performance in \textit{Unseen} set through an inhomogeneous backbone ensemble under IoU $>0.5$ evaluation. \\

\noindent \textbf{Visualizations.} We visualized the inference results of each model to confirm the effect of the inhomogeneous backbone ensemble. Figure 7 shows the visualization of the inference bounding box of the FCOS model with different backbones. As shown in Figure 7, the detectors have the same structure, but only different backbones can be used to obtain very different types of inference results. Through this, it was confirmed that the ensemble model can achieve a very large performance improvement compared to a single model. \\

\begin{figure}[t!]
\centering
\includegraphics[width=1.0\linewidth]{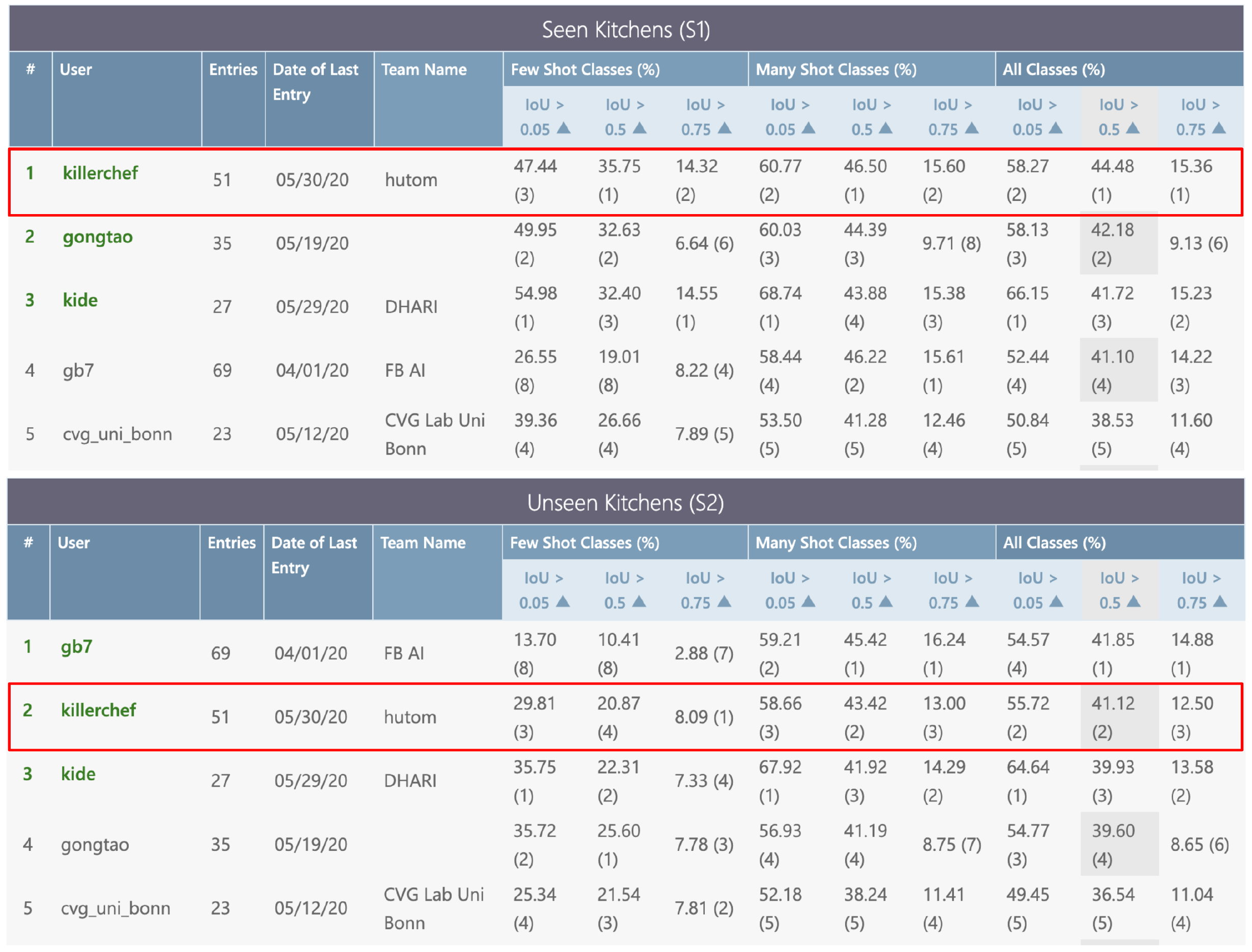}
\caption{\textbf{Epic-Kitchens 2020 object detection challenge evaluation page.} The entry marked with a red box is the final performance evaluated by the our proposed approach. Each entry is ranked under IoU $>0.5$ evaluation.}
\label{fig:example}
\end{figure}

\begin{figure*}[t!]
\centering
\includegraphics[width=1.0\linewidth]{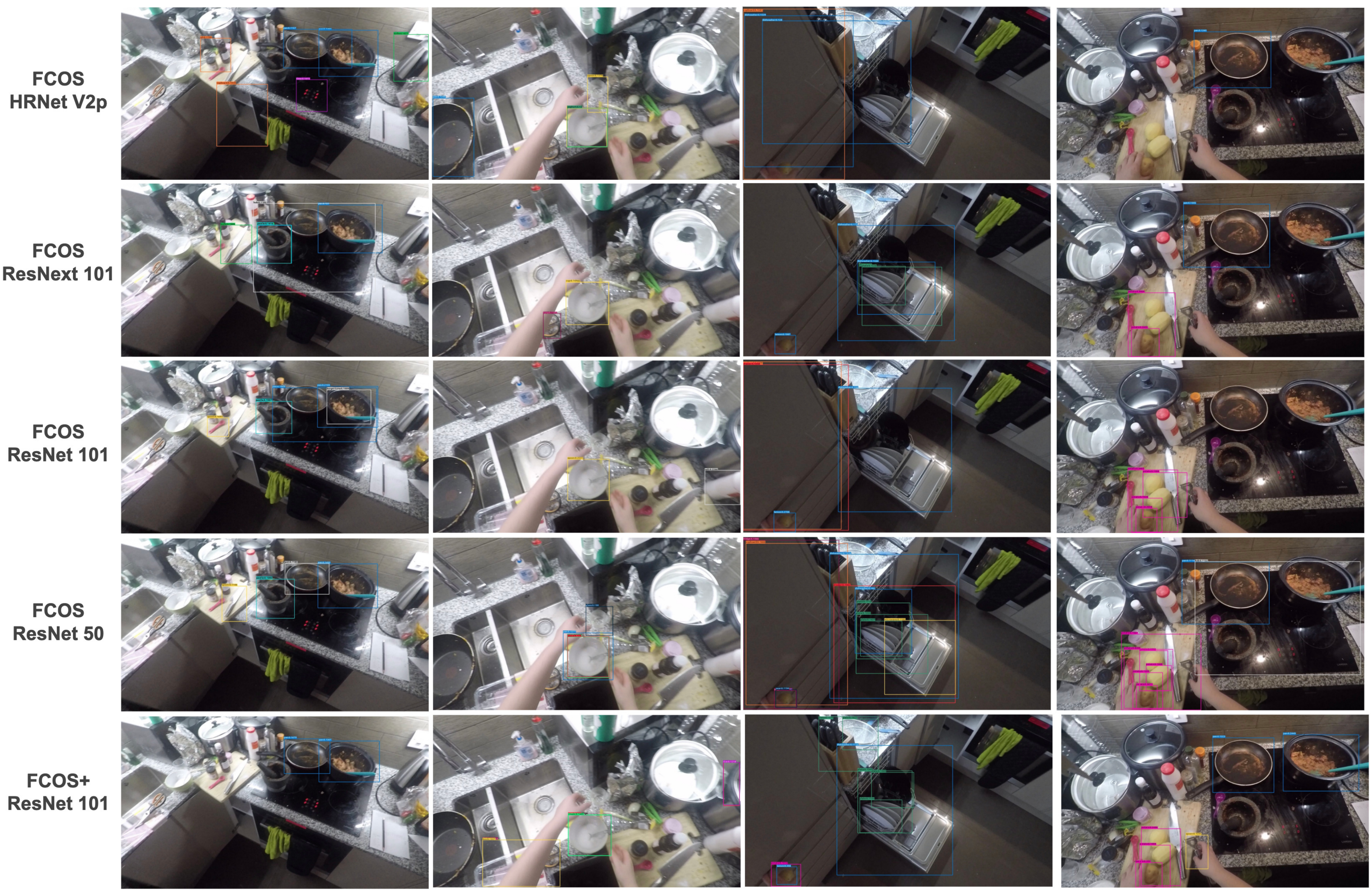}
\caption{\textbf{Visualization for each network prediction.} It shows that the prediction scores can be variously distributed for each network when they have the same structure but have a heterogeneous backbone. The predicted score threshold for visualization was set at 0.5.}
\label{fig:example}
\end{figure*}

\vspace{-3mm}
\section{Conclusion}

We performed single object tracker-based semi-supervised object detection to effectively train datasets with sparse annotations on sequence images. The Epic-Kitchens object detection dataset was used to verify the utility of the proposed technique, and the proposed semi-supervised learning showed good performance in the ensemble as well as in the single model. However, it needs to be analyzed more closely with semi-supervised learning about the advantages and disadvantages of the anchor-based model, and there is a limitation that a simple rule-based engine is used to obtain a pseudo label. For future improvement, it is necessary to perform quantitative analysis on the effect of anchor and RPN on sparse annotation data training, and at the same time, it is possible to consider how to improve the tracking rules or utilize the results obtained in the tracking process during training.

{\small
\bibliographystyle{ieee_fullname}
\bibliography{egbib}

\begin{thebibliography}{10}\itemsep=-1pt

\bibitem{Luca16}
Luca Bertinetto, Jack Valmadre, João~F. Henriques, Andrea Vedaldi, and Philip
  H.~S. Torr.
\newblock Fully-convolutional siamese networks for object tracking.
\newblock In {\em In Proc. of ECCV}, 2016.

\bibitem{Zhaowei18}
Zhaowei Cai and Nuno Vasconcelos.
\newblock Cascade r-cnn: Delving into high quality object detection.
\newblock In {\em In Proc. of CVPR}, 2018.

\bibitem{Bao19}
Bao~Xin Chen and John~K. Tsotsos.
\newblock Fast visual object tracking with rotated bounding boxes.
\newblock In {\em In Proc. of ICCVW}, 2019.

\bibitem{Kai19}
Kai Chen, Jiaqi Wang, Jiangmiao Pang, Yuhang Cao, Yu Xiong, Xiaoxiao Li,
  Shuyang Sun, Wansen Feng, Ziwei Liu, Jiarui Xu, Zheng Zhang, Dazhi Cheng,
  Chenchen Zhu, Tianheng Cheng, Qijie Zhao, Buyu Li, Xin Lu, Rui Zhu, Yue Wu,
  Jifeng Dai, Jingdong Wang, Jianping Shi, Wanli Ouyang, Chen~Change Loy, and
  Dahua Lin.
\newblock Mmdetection: Open mmlab detection toolbox and benchmark.
\newblock {\em arXiv:1906.07155}, 2019.

\bibitem{MinKook18}
Min-Kook Choi, Jaehyeong Park, Jihun Jung, Heechul Jung, Jin-Hee Lee, Woong~Jae
  Won, Woo~Young Jung, Jincheol Kim, and Soon Kwon.
\newblock Co-occurrence matrix analysis-based semi-supervised training for
  object detection.
\newblock In {\em In Proc. of ICIP}, 2018.

\bibitem{Jifeng18}
Jifeng Dai, Yi Li, Kaiming He, and Jian Sun.
\newblock R-fcn: Object detection via region-based fully convolutional
  networks.
\newblock In {\em In Proc. of NIPS}, 2016.

\bibitem{Dima18}
Dima Damen, Hazel Doughty, Giovanni~Maria Farinella, Sanja Fidler, Antonino
  Furnari, Evangelos Kazakos, Davide Moltisanti, Jonathan Munro, Toby Perrett,
  Will Price, and Michael Wray.
\newblock Scaling egocentric vision: The epic-kitchens dataset.
\newblock In {\em In Proc. of ECCV}, 2018.

\bibitem{Mark15}
Mark Everingham, S.~M.~Ali Eslami, Luc~Van Gool, Christopher K.~I. Williams,
  John Winn, and Andrew Zisserman.
\newblock The pascal visual object classes challenge: A retrospective.
\newblock {\em International Journal of Computer Vision}, 111:98--136, 2015.

\bibitem{ChengYang17}
Cheng-Yang Fu, Wei Liu, Ananth Ranga, Ambrish Tyagi, and Alexander~C. Berg.
\newblock Dssd : Deconvolutional single shot detector.
\newblock {\em arXiv:1701.06659}, 2017.

\bibitem{Kaiming16}
Kaiming He, Xiangyu Zhang, Shaoqing Ren, and Jian Sun.
\newblock Deep residual learning for image recognition.
\newblock In {\em In Proc. of CVPR}, 2016.

\bibitem{Heechul17}
Heechul Jung, Min-Kook Choi, Jihun Jung, Jin-Hee Lee, Soon Kwon, and Woo~Young
  Jung.
\newblock Resnet-based vehicle classification and localization in traffic
  surveillance systems.
\newblock In {\em In Proc. of CVPR}, 2017.

\bibitem{Bo19}
Bo Li, Wei Wu, Qiang Wang, Fangyi Zhang, Junliang Xing, and Junjie Yan.
\newblock Siamrpn++: Evolution of siamese visual tracking with very deep
  networks.
\newblock In {\em In Proc. of CVPR}, 2019.

\bibitem{Bo18}
Bo Li, Junjie Yan, Wei Wu, Zheng Zhu, and Xiaolin Hu.
\newblock High performance visual tracking with siamese region proposal
  network.
\newblock In {\em In Proc. of CVPR}, 2018.

\bibitem{TsungYi17}
Tsung-Yi Lin, Priya Goyal, Ross Girshick, Kaiming He, and Piotr Dollár.
\newblock Focal loss for dense object detection.
\newblock In {\em In Proc. of ICCV}, 2017.

\bibitem{TsungYi14}
Tsung-Yi Lin, Michael Maire, Serge Belongie, James Hays, Pietro PeronaDeva,
  Ramanan, Piotr Dollár, and C.~Lawrence Zitnick.
\newblock Microsoft coco: Common objects in context.
\newblock In {\em In Proc. of ECCV}, 2014.

\bibitem{Li18}
Li Liu, Wanli Ouyang, Xiaogang Wang, Paul Fieguth, Jie Chen, Xinwang Liu, and
  Matti Pietikäinen.
\newblock Deep learning for generic object detection: A survey.
\newblock {\em arXiv:1809.02165}, 2018.

\bibitem{Wei16}
Wei Liu, Dragomir Anguelov, Dumitru Erhan, Christian Szegedy, Scott Reed,
  Cheng-Yang Fu, and Alexander~C. Berg.
\newblock Ssd: Single shot multibox detector.
\newblock In {\em In Proc. of ECCV}, 2016.

\bibitem{Ishan15}
Ishan Misra, Abhinav Shrivastava, and Martial Hebert.
\newblock Watch and learn: Semi-supervised learning for object detectors from
  video.
\newblock In {\em In Proc. of CVPR}, 2015.

\bibitem{Yusuke19}
Yusuke Niitani, Takuya Akiba, Tommi Kerola, Toru Ogawa, Shotaro Sano, and Shuji
  Suzuki.
\newblock Sampling techniques for large-scale object detection from sparsely
  annotated objects.
\newblock In {\em In Proc. of CVPR}, 2019.

\bibitem{Jordi17}
Jordi Pont-Tuset, Federico Perazzi, Sergi Caelles, Pablo Arbelaez, Alexander
  Sorkine-Hornung, and Luc~Van Gool.
\newblock The 2017 davis challenge on video object segmentation.
\newblock {\em arXiv:1704.00675}, 2017.

\bibitem{Joseph17}
Joseph Redmon and Ali Farhadi.
\newblock Yolo9000: Better, faster, stronger.
\newblock In {\em In Proc. of CVPR}, 2017.

\bibitem{Joseph18}
Joseph Redmon and Ali Farhadi.
\newblock Yolov3: An incremental improvement.
\newblock {\em arXiv:1804.02767}, 2018.

\bibitem{Shaoqing15}
Shaoqing Ren, Kaiming He, Ross Girshick, and Jian Sun.
\newblock Faster r-cnn: Towards real-time object detection with region proposal
  networks.
\newblock In {\em In Proc. of NIPS}, 2015.

\bibitem{Zhi19}
Zhi Tian, Chunhua Shen, Hao Chen, and Tong He.
\newblock Fcos: Fully convolutional one-stage object detection.
\newblock In {\em In Proc. of ICCV}, 2019.

\bibitem{Thang19}
Thang Vu, Hyunjun Jang, Trung~X. Pham, and Chang Yoo.
\newblock Cascade rpn: Delving into high-quality region proposal network with
  adaptive convolution.
\newblock In {\em In Proc. of NIPS}, 2019.

\bibitem{Jingdong19}
Jingdong Wang, Ke Sun, Tianheng Cheng, Borui Jiang, Chaorui Deng, Yang Zhao,
  Dong Liu, Yadong Mu, Mingkui Tan, Xinggang Wang, Wenyu Liu, and Bin Xiao.
\newblock Deep high-resolution representation learning for visual recognition.
\newblock {\em arXiv:1908.07919}, 2019.

\bibitem{Qiang19}
Qiang Wang, Li Zhang, Luca Bertinetto, Weiming Hu, and Philip~H.S. Torr.
\newblock Fast online object tracking and segmentation: A unifying approach.
\newblock In {\em In Proc. of CVPR}, 2019.

\bibitem{Saining17}
Saining Xie, Ross Girshick, Piotr Dollar, Zhuowen Tu, and Kaiming He.
\newblock Aggregated residual transformations for deep neural networks.
\newblock In {\em In Proc. of CVPR}, 2017.

\bibitem{Shifeng18}
Shifeng Zhang, Longyin Wen, Xiao Bian, Zhen Lei, and Stan~Z. Li.
\newblock Single-shot refinement neural network for object detection.
\newblock In {\em In Proc. of CVPR}, 2018.

\end{thebibliography}
}

\end{document}